\documentclass{article}

\PassOptionsToPackage{numbers}{natbib}

\usepackage[preprint]{main}

\usepackage[utf8]{inputenc} 
\usepackage[T1]{fontenc}    
\usepackage{hyperref}       
\usepackage{url}            
\usepackage{booktabs}       
\usepackage{amsfonts}       
\usepackage{nicefrac}       
\usepackage{microtype}      
\usepackage{xcolor}         
\usepackage{wrapfig}
\usepackage{amsmath}
\usepackage{tabularx}
\usepackage{color}
\usepackage{subcaption}
\usepackage[justification=raggedright,singlelinecheck=false]{caption}
\usepackage{comment}

\usepackage{multirow}
\usepackage{graphicx} 
\usepackage{booktabs}
\usepackage{array}

\newcolumntype{C}{>{\centering\arraybackslash}X}

\definecolor{green}{RGB}{78,166,46}
\colorlet{green50}{green!10} 
\definecolor{blue}{RGB}{15,158,213}
\colorlet{blue50}{blue!10}
\definecolor{orange}{RGB}{233,113,5}
\colorlet{orange50}{orange!10} 

\title{Dynamic Pattern Alignment Learning for Pretraining Lightweight Human-Centric Vision Models}

\author{%
	Xuanhan Wang$^{1}$ \quad \quad  \quad Huimin Deng$^{2}$   \quad \quad \quad Ke Liu$^2$ \quad \quad \quad  Jun Wang$^2$ \\
	\textbf{Lianli Gao}$^2$ \quad  \quad \quad \textbf{Jingkuan Song}$^1$ \\
	\\
	$^1$Tongji University\\ \\
	$^2$University of Electronic Science and Technology of China  \\
}

\begin{document}
\maketitle

\begin{abstract}
    Human-centric vision models (HVMs) have achieved remarkable generalization due to large-scale pretraining on massive person images. However, their dependence on large neural architectures and the restricted accessibility of pretraining data significantly limits their practicality in real-world applications. To address this limitation, we propose Dynamic Pattern Alignment Learning (DPAL), a novel distillation-based pretraining framework that efficiently trains lightweight HVMs to acquire strong generalization from large HVMs. 
	In particular, human-centric visual perception are highly dependent on three typical visual patterns, including global identity pattern, local shape pattern and multi-person interaction pattern. 
	To achieve generalizable lightweight HVMs, we firstly design a dynamic pattern decoder (D-PaDe), acting as a dynamic Mixture of Expert (MoE) model. It incorporates three specialized experts dedicated to adaptively extract typical visual patterns, conditioned on both input image and pattern queries.
	And then, we present three levels of alignment objectives, which aims to minimize generalization gap between lightweight HVMs and large HVMs at global image level, local pixel level, and instance relation level. 
	With these two deliberate designs, the DPAL effectively guides lightweight model to learn all typical human visual patterns from large HVMs, which can generalize to various human-centric vision tasks. 
	Extensive experiments conducted on 15 challenging datasets demonstrate the effectiveness of the DPAL. Remarkably, when employing PATH-B as the teacher, DPAL-ViT/Ti (5M parameters) achieves surprising generalizability similar to existing large HVMs such as PATH-B (84M) and Sapiens-L (307M), and outperforms previous distillation-based pretraining methods including Proteus-ViT/Ti (5M) and TinyMiM-ViT/Ti (5M) by a large margin. More importantly, the DPAL is performed on a limited dataset (i.e., around 1M unlabeled images) that is unseen for large HVMs, which bypasses the need for those inaccessible or constrained pretraining datasets, offering an affordable approach to generalizable HVMs. All code and checkpoints will be publicly available\footnote{\url{https://github.com/Hermine-D/DPAL}}.
\end{abstract}

\section{Introduction}
\begin{figure}[h]
  \centering
  \includegraphics[width=\linewidth]{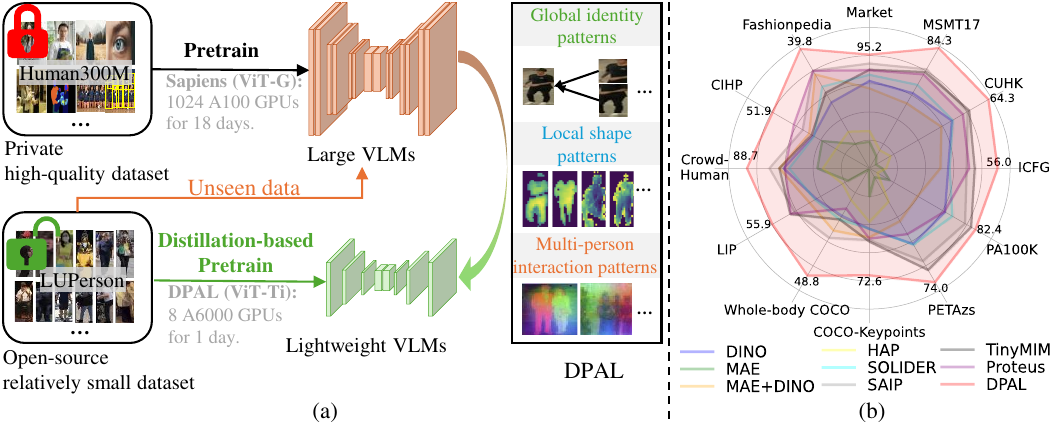}
  \caption{(a) Existing HVMs are limited in real-world application due to two factors: 1) Large model size with heavy computational costs, and 2) high-quality pretraining datasets are strictly constrained. To address these, Dynamic Pattern Alignment Learning (DPAL) is proposed to pretrain lightweight HVMs by distilling generalization capability from large HVMs across three typical human visual patterns. (b) Extensive experiments conducted on 12 datasets demonstrate effectiveness of the DPAL.}
  \label{fig:motivation}
\end{figure}

Recent years have witnessed remarkable progresses in human-centric visual perception (HVP)~\cite{yuan2023hap,tang2023humanbench,ci2023unihcp,khirodkar2024sapiens,solider}. This success is mainly attributed to the advancement in pretraining of large vision models with massive collected data. By leveraging such extensive pretraining, large human-centric vision models (HVMs) are able to learn generalizable visual patterns, which widely benefit various human-centric perception tasks, such as single-person discrimination~\cite{fu2021unsupervised,reid2}, dense prediction~\cite{pose3,pose4} and multi-person visual understanding~\cite{ci2023unihcp,tang2023humanbench}. 

Although large HVMs exhibit strong generalization capability, there are two primary computational challenges that significantly limit the practicality of large HVMs in real-world application. 
First, large HVMs typically exhibit substantial model size, demanding considerable computational resources and making the pretraining of HVMs prohibitively expensive for most researchers. For example, Sapiens~\cite{khirodkar2024sapiens}, a typical self-supervised pretrained HVM, employs ViT-G (2B parameters) as the model architecture and trains it on Humans-300M for 18 days using 1024 A100 GPUs. The PATH, a representative supervised pretrained HVM, trains ViT-L on HumanBench with 11M images using 64 V100 GPUs. This substantial dependence on huge computational resources complicates real-world deployment of HVMs, resulting in slow inference speed especially on resource-constrained devices.
Second, pretraining HVMs relies on extensive and high-quality datasets. However, the qualified pretraining datasets such as Humans-300M~\cite{khirodkar2024sapiens} and HumanBench~\cite{tang2023humanbench} are often inaccessible or strictly constrained due to the concern about violations to copyright ownership of these digital assets. These challenges pose significant limitation on the broad applicability of existing pretraining methods. 

To achieve strong generalizability while maintain the broad applicability, exploiting the potential of lightweight HVMs via extensive pretraining is a promising direction. Previous works~\cite{dino,he2022masked,solider,yuan2023hap} on lightweight HVMs focus on self-supervised pretraining by using alternative dataset such as LUPerson~\cite{fu2021unsupervised}, which has long been the widely used benchmark for advancements in human-centric vision foundation models. Nevertheless, the scale of publicly available dataset is relatively small, e.g., the number of image in LUPerson is around 4M which is much less than that of those private datasets, limiting the optimization of lightweight HVMs. Motivated by these issues, we focus on a significant question that is considerably less studied: \textbf{\textit{is it possible to replicate strong generalization capability from large HVMs to lightweight HVMs without requiring access to those inaccessible or strictly constrained pretraining datasets?}} 

In this work, we seek to answer the question by exploring distillation-based pretraining framework, which trains lightweight HVMs to acquire strong generalization from large HVMs via limited dataset. In general, human-centric visual perception relies on three typical visual patterns: 1) global identity pattern for single-person discrimination tasks such as Person ReID or attribute recognition, 2) local shape pattern for dense prediction tasks such as pose estimation or human parsing, and 3) multi-person interaction pattern for multi-person visual understanding tasks like pedestrian detection or part-level attribute parsing. This reliance suggests that a robust human-centric vision model with strong generalization must be capable of acquiring all three visual patterns. It is should be noted that directly learning different patterns is hindered by inter-pattern conflict, inevitably yielding suboptimal visual representations. As analyzed in previous works~\cite{solider, yuan2023hap}, global pattern learning tends to homogenize pixel representations, thereby sacrificing fine-grained information. However, local pattern learning hurts the global identity information, as it is expected to learn semantic-consistent region representations. Motivated by these, we propose \textbf{D}ynamic \textbf{P}attern \textbf{A}lignment \textbf{L}earning (DPAL), a novel pretraining distillation framework that dynamically transfers learned diverse patterns from large HVMs to lightweight HVMs. 
Specifically, we firstly design a dynamic pattern decoder (D-PaDe), acting as a dynamic MoE model with three experts dedicated to separately processing local, global, and relational patterns. It activates only one expert module per input to ensure alignment with one pattern.
And then, we present three levels of alignment objectives, which aims to minimize generalization gap between lightweight HVMs and large HVMs at global image level, local pixel level, and instance relation level.
With these deliberate designs, the DPAL effectively alleviates inter-pattern conflict and successfully guides lightweight HVMs to learn all typical human visual patterns, thereby acquiring strong generalization capability from large HVMs.

We conduct extensive experiments on 15 challenging benchmarks involving 9 representative human-centric visual perception tasks and three cross-domain visual perception tasks, demonstrating the impressive effectiveness of proposed DPAL.
Remarkably, when employing PATH-B as the teacher, DPAL-ViT/Ti (5M parameters) achieves surprising generalizability similar to existing large HVMs such as PATH-B (84M) and Sapiens-L (307M), and outperforms distillation-based pretraining methods including Proteus-ViT/Ti (5M) and TinyMiM-ViT/Ti (5M) by a large margin. More importantly, the distillation pretraining in DPAL is performed on the limited dataset that is unseen for large HVMs for around 1 day, without the need for those inaccessible or constrained pretraining datasets, offering an affordable approach to generalizable HVMs.

\section{Related Works}
\subsection{Human-centric Vision Model}
Human-centric vision models (HVMs) refer to pretrained models specifically designed and trained to handle human-related visual tasks, including person re-identification (ReID) \cite{reid1, reid2, reid3}, text-to-image person ReID \cite{t2ireid1, t2ireid2, t2ireid3, t2ireid4}, pedestrian attribute recognition \cite{attribute1, attribute2}, action recognition \cite{action1, action2, action3}, and body structure understanding such as 2D/3D pose estimation \cite{pose1, pose2, pose3, pose4}. 

Recent works \cite{yuan2023hap, ci2023unihcp, tang2023humanbench, khirodkar2024sapiens, solider} have proposed HVMs tailored for human-centric tasks.
HAP ~\cite{yuan2023hap} introduces human body structure priors into the pretraining task, encouraging the model to focus on human body structure information, and has achieved significant results in 11 human-centered benchmark tests. 
PATH ~\cite{tang2023humanbench} proposes a hierarchical weight-sharing strategy and performs multi-task training on five tasks, enabling the model to extract both common features across tasks and task-specific features. 
Unihcp ~\cite{ci2023unihcp} further increases the parameter sharing of the backbone across different tasks, using only Task-Specific Queries to trigger the model's output for different downstream tasks. 
Sapiens ~\cite{khirodkar2024sapiens} collected a large number of high-quality in-the-wild human images, Humans-300M, and pre-trained the model on this dataset, followed by MAE. The model was then validated on four human-centric downstream tasks.

Although the above methods have achieved impressive results in human-centric downstream tasks, the large parameter size and inaccessible large-scale dataset of HVMs \cite{yuan2023hap, tang2023humanbench, ci2023unihcp, khirodkar2024sapiens} makes them unsuitable for real-world application. Therefore, our primary objective is to leverage knowledge distillation to effectively transfer human-centric representations from large HVMs to a lightweight counterpart without obtain source dataset while preserving strong generalization capability.

\subsection{Knowledge Distillation}
Knowledge distillation ~\cite{hinton2015distilling, distill_2, distill_3, distill_4, vitkd, maskedkd, scalekd} compresses the knowledge from one or more larger models into a smaller model. To leverage visual foundation models, which are usually computationally intensive, several studies have explored distilling them into more compact models.

DeiT ~\cite{deit} addresses the issue of vision transformer training requiring a large labeled dataset by jointly using a class token and a distillation token during training on ImageNet~\cite{deng2009imagenet}.
TinyMIM ~\cite{tinymim} explores the impact of different distillation methods using an masked image modeling~(MIM) pre-trained model, thereby extending the capabilities of MIM models to smaller models.
Unlike traditional single-stage distillation, G2SD ~\cite{g2sd} proposes a two-stage distillation approach, where a MAE pre-trained model is used as the teacher model for distillation, followed by distillation on specific tasks.
Furthermore, theia ~\cite{theia}  proposes a multi-teacher off-the-shelf distillation model, which learns from the rich visual representations of multiple teachers simultaneously.
Proteus ~\cite{proteus} conduct distillation across three different levels of training objectives to transfer the fruitful knowledge by emulating the teacher’s behaviors.

These methods illustrate the considerable potential of knowledge distillation in model compression and efficiency improvement. However, existing methods are limited to a single visual pattern (e.g. DeiT\cite{deit} captures global identity patterns, G2SD\cite{g2sd} focuses on local patterns), resulting in a significant gap in human-centric visual perception tasks. To overcome this limitation, we introduce DPAL, which dynamically learns three typical patterns from large HVMs to enhance generalizability across diverse downstream tasks. 

\section{Dynamic Pattern Alignment Learning} \label{dpal}
\begin{figure}[h]
  \centering
  \includegraphics[width=\linewidth]{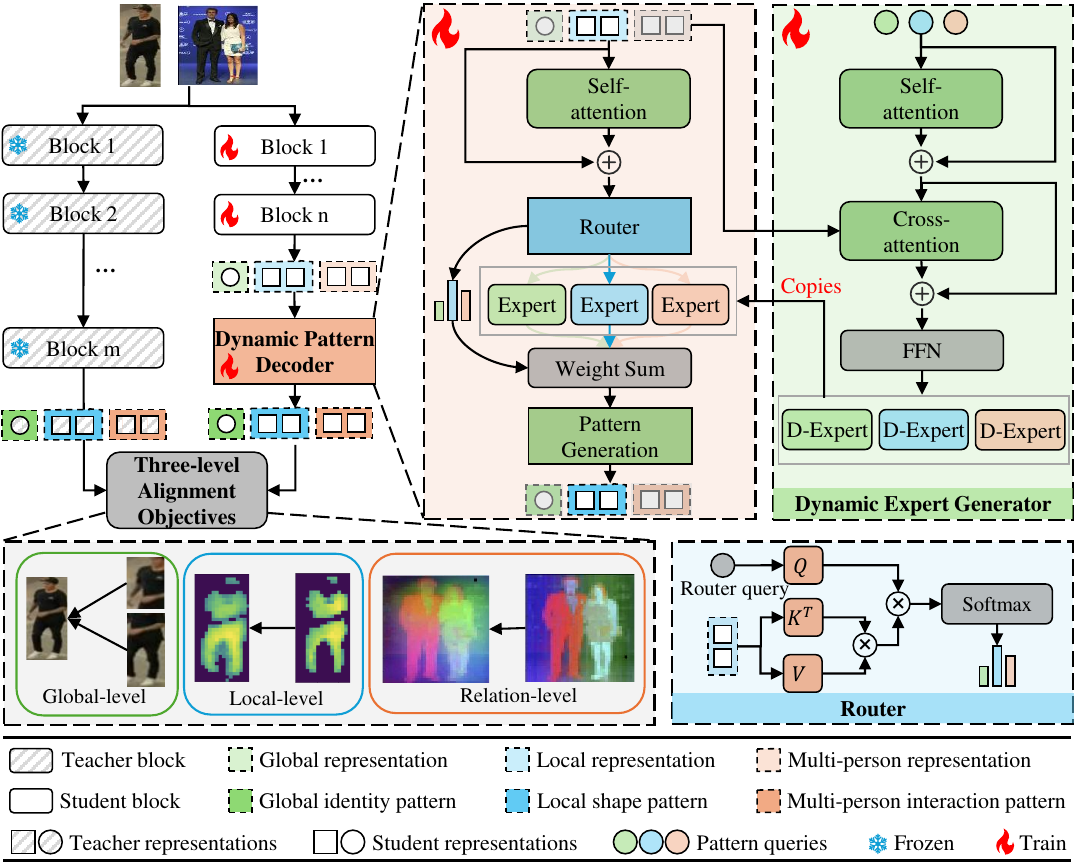}
  \caption{The overview of DPAL. It comprises of (1) Teacher model, (2) Student backbone and (3) Dynamic pattern decoder. Dynamic pattern decoder captures three types of patterns conditioned on the input image and pattern queries. Finally, three-level alignment objectives are employed to acquire generalization capability from large HVMs to lightweight HVMs.
  }
  \label{fig:framework}
\end{figure}

As illustrated in Fig.~\ref{fig:framework}, we propose \textbf{D}ynamic \textbf{P}attern \textbf{A}lignment \textbf{L}earning (DPAL), a distillation-based pretraining framework for building generalizable lightweight HVMs. Following standard teacher-student architecture~\cite{dino,g2sd,hinton2015distilling,proteus,theia}, the DPAL relies on two major designs: 1) Dynamic Pattern Decoder that extracts three typical human visual patterns in a dynamic way, and 2) three levels of alignment objectives that maximize knowledge transfer from the large HVMs to the lightweight HVMs through these visual patterns. 

\subsection{Architecture Design of DPAL} \label{Preliminaries}
Similar to standard knowledge distillation framework, our distillation-based pretraining framework consists of three major components: visual encoder of student model (s-VisEn), visual encoder of teacher model (t-VisEn) and Dynamic Pattern Decoder (D-PaDe). After pretraining, only s-VisEn is retained for downstream tasks while the t-VisEn and D-PaDe are discarded.

\subsubsection{Basic Visual Encoder}
\textbf{s-VisEn.} The basic architecture of s-VisEn is built on lightweight vision transformer (e.g., ViT/Ti), which tokenizes input image into numerous vision tokens. Formally, given an input image \( I\in\mathbb{R}^{C\times H \times W} \), it outputs two types of tokens, i.e., global image token \( F_s^{g}\in\mathbb{R}^{D_s} \) representing $D_s$ dimensional global representation of an image, and local patch tokens \( F_s^{l}\in\mathbb{R}^{L \times D_s} \) representing local representations of $L$ image regions.

\textbf{t-VisEn.} In general, the teacher model is a pretrained large vision transformer, which has a deeper or wider representation capacity for providing diverse visual patterns. In this work, the t-VisEn is utilized to extract global image tokens \( F_t^{g}\in\mathbb{R}^{D_t} \), local patch tokens \( F_t^{l}\in\mathbb{R}^{L\times D_t} \) and attention tokens \( F_t^{r}\in\mathbb{R}^{L\times L} \), which implicitly represent three typical human vision patterns.



\subsubsection{Dynamic Pattern Decoder } 

To well acquire diverse visual patterns from the teacher model, the D-PaDe is designed as a dynamic mixture of experts (MoE) model, which adaptively casts input tokens as the typical visual patterns. As illustrated in Fig.~\ref{fig:framework}, the core design of D-PaDe relies on dynamic expert generator. 
Unlike common MoE model that pre-defines a set of fix computational experts, the computational experts in proposed dynamic expert generator are dynamically predicted based on the input visual tokens and pattern queries. Specifically, the pattern queries are defined as three learnable tokens \(T_e=[T_e^1, T_e^2,T_e^3]\), which subsequently pass through self-attention, cross-attention, and FFN modules to predict the parameters of three computational experts $\{E\}^3_{i=1}$. In particular, each of experts is a small feed-forward network (FFN) with linear transformation parameters. Using the visual tokens from s-VisEn as the keys and values, the cross-attention module enables predicted parameters to focus on one specific visual pattern of the input image. In the following, we present how to utilize D-PaDe to extract typical visual patterns.

\textbf{Global identity pattern} is generally expressed via global discrimination information. Therefore, this pattern can be directly obtained from single-person image $I_1$, as formulated as follows:
\begin{equation}
	\widetilde{F_s^g}=\sum^{3}_{i=1}W_{i}^e\cdot E_i(F_{s1}^{g}+SA(F_{s1}^{g}))
\end{equation}
where $F_{s1}^{g}$ is the global token from s-VisEn for the single-person image, and $SA(\cdot)$ represents self-attention. The $W_{i}^e$ is the weight score for $i$-th expert.

\textbf{Local shape pattern} represents local human body shape information. To achieve this shape information, we use the t-VisEn to extract a coarse mask $M_{shape}$ that attends foreground body shape. Then, extracted mask is used to filter local patch tokens for local shape pattern generation. This process can be formulated in Equ.~\ref{equ.local_shape}:
\begin{equation}
	\widetilde{F_{s}^{l}}=M_{shape}\cdot \left[ \sum^{3}_{i=1}W_{i}^e\cdot E_i(F_{s1}^{l}+SA(F_{s1}^{l}))\right]
	\label{equ.local_shape}
\end{equation}
where $F_{s1}^{l}$ is the local patch token from s-VisEn for the single-person image.

\textbf{Multi-person interaction pattern} represents the relational information between different instances. To generate this kind of pattern, we firstly extract image patch tokens $F_{s2}^{l}$ from a multi-person image $I_2$ via the s-VisEn, and then input these tokens into the D-PaDe for multi-person feature extraction. After that, a self-attention module is used to predict the correlation scores among the patches in multi-person feature. This process is formulated through Equ.~\ref{equ.multi_pattern}:
\begin{equation}
\begin{array}{lll}
		\widetilde{F_{s2}^{l}} &=& \sum^{3}_{i=1}W_{i}^e\cdot E_i(F_{s2}^{l}+SA(F_{s2}^{l}))  \vspace{0.5em}\\ 
		\widetilde{F_{s}^{r}} &=& softmax(\frac{\widetilde{F_{s2}^{l}}^T\widetilde{F_{s2}^{l}}}{\sqrt{D_s}}) 
\end{array}
	\label{equ.multi_pattern}
\end{equation}

\subsection{Three Levels of Alignment Objectives} \label{Multi-level Learning Objectives}
To minimize the generalization gap between lightweight HVMs and large HVMs, we conduct pattern alignment across three different levels, i.e., global-level, local-level, and relation-level.

\textbf{Global-level Alignment.} To learn global identity pattern, we construct \(M\) multi-view images derived from the single-person image $I_1$. Then, we extract global identity patterns from those multi-view images and minimize their distribution gap between the student model and teacher model, using MSE loss as formulated in Equ \ref{eq:cscl}.
\begin{equation}
	\ell_{g} = \frac{1}{M}\sum_{i=1}^{M} \Vert\widetilde{F_s^g}_i-\widetilde{F_t^g}\Vert_2
	\label{eq:cscl}
\end{equation}

\textbf{Local-level Alignment.} To align local shape pattern, we use the MSE loss to encourage the consistency between patch tokens decoded via proposed D-PaDe and local shape pattern extracted from the t-VisEn, as illustrated in Equ \ref{eq:cspl}:
\begin{equation}
	\ell_{l} = \Vert\widetilde{F_s^l}-\widetilde{F_t^l}\Vert_2
	\label{eq:cspl}
\end{equation}

\textbf{Relation-level Alignment.} To further acquire instance-level relationships learned in the teacher model, we enforce the lightweight HVMs mimic the multi-person interaction patterns of large HVMs via KL divergence loss:
\begin{equation}
	\ell_{r} = L_{KL}(\widetilde{F_{s}^{r}},\widetilde{F_{t}^{r}})
	\label{eq:csal}
\end{equation}

The overall learning objective $\mathcal{L}$ in DPAL can be expressed as follows in Equ~\ref{equ.dpal_loss}:
\begin{equation}
	\begin{array}{lll}
		& \mathop{min}\limits_{\theta} \mathbb{E}[\mathcal{L}(I_1, I_2)],  & \\ 
		& s.t., \,\,\, \mathcal{L}=\ell_{g}+\ell_{l}+\ell_{r} &
	\end{array}
	\label{equ.dpal_loss} 
\end{equation}
where $\theta$ is the learnable parameters of a lightweight HVMs.

\section{Experiment} \label{Experiment}
\begin{table}[ht!]
    \setlength{\tabcolsep}{1.5pt}
    \renewcommand\arraystretch{1.25}
    \caption{Comparison with self-supervised pretraining and distillation-based pretraining methods across single-person discriminative tasks (a)(b)(c), and single-person dense prediction tasks (d)(e)(f). $^\star$: Swin-tiny is adopted as the student backbone.} 
    \label{tab:comparison with sota}
    { 
      \fontsize{9pt}{10pt}\selectfont
      \setlength{\tabcolsep}{2pt}  
      \renewcommand{\arraystretch}{1.2}  
    \begin{subtable}{.32\textwidth}
    \caption{I2I Person ReID (\%).} \label{table:i2ireid}
    \centering
    \begin{tabularx}{\linewidth}{m{0.38\linewidth} >{\centering\arraybackslash}m{0.27\linewidth} >{\centering\arraybackslash}m{0.23\linewidth}}
        \hline
        Method & Market$\uparrow$ & MSMT17$\uparrow$ \\
        \hline
        DINO\cite{dino}         & 90.5 & 65.8 \\
        MAE\cite{he2022masked}  & 79.7 & 39.9 \\
        MAE+DINO\cite{mae+dino} & 89.2 & 61.6 \\
        HAP\cite{yuan2023hap}   & 81.6 & 42.4 \\
        SOLIDER\cite{solider}   & 91.6 & 69.2 \\
        SAIP\cite{saip}         & 93.6 & 75.6 \\
        \hline
        ViTKD\cite{vitkd}   & 88.3 & 65.7 \\
        MaskedKD\cite{maskedkd}   & 87.9  & 62.1 \\
        ScaleKD\cite{scalekd}   & 90.7 & 68.9 \\
        TinyMIM\cite{tinymim}   & 92.5 & 74.5 \\
        Proteus\cite{proteus}   & 92.4 & 73.5 \\
        \hline
        DPAL	 & \underline{95.2} & \underline{84.3} \\
        DPAL$^{\star}$ & \textbf{96.4} & \textbf{86.2} \\
        \hline
    \end{tabularx}
    \end{subtable}
    ~
    \begin{subtable}{.32\textwidth}
    \caption{T2I Person ReID (\%).} \label{table:t2ireid}
    \begin{tabularx}{\linewidth}{m{0.38\linewidth} >{\centering\arraybackslash}m{0.30\linewidth} >{\centering\arraybackslash}m{0.23\linewidth}}
        \hline
        Method & CUHK$\uparrow$ & ICFG$\uparrow$ \\
        \hline
        DINO\cite{dino}         & 55.3 & 40.2 \\
        MAE\cite{he2022masked}  & 36.6 & 19.1 \\
        MAE+DINO\cite{mae+dino} & 52.7 & 37.7 \\
        HAP\cite{yuan2023hap}   & 40.2 & 20.4 \\
        SOLIDER\cite{solider}   & 55.5 & 40.7 \\
        SAIP\cite{saip}         & 59.2 & 46.1 \\
        \hline
        ViTKD\cite{vitkd}   & 46.3 & 35.5 \\
        MaskedKD\cite{maskedkd}   & 52.4 & 42.8 \\
        ScaleKD\cite{scalekd}   & 57.4 & 46.5 \\
        TinyMIM\cite{tinymim}   & 59.6 & 48.7 \\
        Proteus\cite{proteus}   & 58.0 & 46.9 \\
        \hline
        DPAL	 & \underline{64.3} & \underline{56.0} \\
        DPAL$^{\star}$ & \textbf{66.9} & \textbf{58.5} \\
        \hline
    \end{tabularx}
    \end{subtable}
    ~
    \begin{subtable}{.33\textwidth}
    \caption{Attribute Recognition (\%).} \label{table:attribute}
    \begin{tabularx}{\linewidth}{m{0.38\linewidth} >{\centering\arraybackslash}m{0.30\linewidth} >{\centering\arraybackslash}m{0.23\linewidth}}
        \hline
        Method & PA-100K$\uparrow$ &  PETAzs$\uparrow$ \\
        \hline
        DINO\cite{dino}         & 77.4 & 69.3 \\
        MAE\cite{he2022masked}  & 68.3 & 61.1 \\
        MAE+DINO\cite{mae+dino} & 72.9 & 66.5 \\
        HAP\cite{yuan2023hap}   & 66.3 & 64.1 \\
        SOLIDER\cite{solider}   & 78.6 & 69.4 \\
        SAIP\cite{saip}         & 80.7 & 71.4 \\
        \hline
        ViTKD\cite{vitkd}   & 77.6 &  67.0\\
        MaskedKD\cite{maskedkd}   & 75.6 & 67.6 \\
        ScaleKD\cite{scalekd}   & 78.2 & 69.3 \\
        TinyMIM\cite{tinymim}   & 81.7 & 72.5 \\
        Proteus\cite{proteus}   & 77.3 & 68.1 \\
        \hline
        DPAL	 & \underline{82.4} & \underline{74.0} \\
        DPAL$^{\star}$ & \textbf{83.1} & \textbf{74.9} \\
        \hline
     \end{tabularx}
     \end{subtable}
    \vspace{1mm}
    \begin{subtable}[t]{.32\textwidth}
    \caption{Pose Estimation (\%).}\label{table:pose}
    \begin{tabularx}{\linewidth}{m{0.38\linewidth} >{\centering\arraybackslash}m{0.30\linewidth} >{\centering\arraybackslash}m{0.23\linewidth}}
        \hline
        Method   & $AP\uparrow$ & $AR\uparrow$ \\
        \hline
        DINO\cite{dino}         & 69.3 & 72.6 \\
        MAE\cite{he2022masked}  & 67.0 & 70.6 \\
        MAE+DINO\cite{mae+dino} & 69.9 & 73.2 \\
        HAP\cite{yuan2023hap}   & 68.8 & 72.3 \\
        SOLIDER\cite{solider}   & 69.3 & 72.6 \\
        SAIP\cite{saip}         & 70.1 & 73.3 \\
        \hline
        ViTKD\cite{vitkd}   & 71.3 & 74.5 \\
        MaskedKD\cite{maskedkd}   & 67.9 & 71.5 \\
        ScaleKD\cite{scalekd}   & 70.8 & 74.1 \\
        TinyMIM\cite{tinymim}   & 70.2 & 73.6 \\
        Proteus\cite{proteus}   & 70.0 & 73.3 \\
        \hline
        DPAL	 & \underline{72.6} & \underline{75.8} \\
        DPAL$^{\star}$ & \textbf{75.1} & \textbf{78.1} \\
        \hline
    \end{tabularx}
    \end{subtable}
    ~
    \begin{subtable}[t]{.32\textwidth}
    \caption{Landmark Detection (\%).}\label{table:landmark}
    \begin{tabularx}{\linewidth}{m{0.38\linewidth} >{\centering\arraybackslash}m{0.30\linewidth} >{\centering\arraybackslash}m{0.23\linewidth}}
        \hline
        Method   & $AP\uparrow$ & $AR\uparrow$ \\
        \hline
        DINO\cite{dino}         & 43.9 & 57.1 \\
        MAE\cite{he2022masked}  & 40.1 & 52.7 \\
        MAE+DINO\cite{mae+dino} & 45.1 & 58.2 \\
        HAP\cite{yuan2023hap}   & 42.6 & 55.4 \\
        SOLIDER\cite{solider}   & 44.2 & 57.2 \\
        SAIP\cite{saip}         & 45.7 & 58.7 \\
        \hline
        ViTKD\cite{vitkd}   & 44.4 & 57.3 \\
        MaskedKD\cite{maskedkd}   & 40.4  & 54.0 \\
        ScaleKD\cite{scalekd}   & 43.8 & 57.0 \\
        TinyMIM\cite{tinymim}   & 44.2 & 57.5 \\
        Proteus\cite{proteus}   & 43.3 & 56.4 \\
        \hline
        DPAL	 & \underline{48.8} & \underline{61.5} \\
        DPAL$^{\star}$ & \textbf{53.9} & \textbf{65.7} \\
        \hline
    \end{tabularx}
    \end{subtable}
    ~
    \begin{subtable}[t]{.32\textwidth}
    \caption{Human Parsing (\%).}\label{table:seg}
    \centering
    \begin{tabularx}{\linewidth}{m{0.38\linewidth} >{\centering\arraybackslash}m{0.30\linewidth} >{\centering\arraybackslash}m{0.23\linewidth}}
        \hline
        Method   & $mIoU\uparrow$ & $mAcc\uparrow$ \\
        \hline
        DINO\cite{dino}         & 48.7 & 59.3 \\
        MAE\cite{he2022masked}  & 43.5 & 54.0 \\
        MAE+DINO\cite{mae+dino} & 49.8 & 60.3 \\
        HAP\cite{yuan2023hap}   & 44.4 & 54.5 \\
        SOLIDER\cite{solider}   & 48.9 & 59.2 \\
        SAIP\cite{saip}         & 52.3 & 63.3 \\
        \hline
        ViTKD\cite{vitkd}   & 52.0 & 62.8 \\
        MaskedKD\cite{maskedkd}   & 50.9 & 61.7  \\
        ScaleKD\cite{scalekd}   & 55.6 & 66.5 \\
        TinyMIM\cite{tinymim}   & 53.0 & 63.7 \\
        Proteus\cite{proteus}   & 52.9 & 63.9 \\
        \hline
        DPAL	 & \underline{55.9} & \underline{66.7} \\
        DPAL$^{\star}$ & \textbf{59.3} & \textbf{69.7} \\
        \hline
    \end{tabularx}
    \end{subtable}
}
\end{table}

\begin{table}[ht!]
	\setlength{\tabcolsep}{1.5pt}
	\renewcommand\arraystretch{1.25}
	\caption{Comparison with state-of-the-arts across three multi-person perception tasks (a)(b)(c), and cross-domain visual perception tasks (d)(e)(f). $^\star$: Swin-tiny is adopted as the student backbone.} 
	\label{tab:comparison with sota_interaction}
	{ 
		\fontsize{9pt}{10pt}\selectfont
		\setlength{\tabcolsep}{2pt}  
		\renewcommand{\arraystretch}{1.2}  

	\vspace{1mm}
	\begin{subtable}[t]{.32\textwidth}
		\caption{Pedestrian Detection (\%).}\label{table:detection}
		\begin{tabularx}{\linewidth}{m{0.38\linewidth} >{\centering\arraybackslash}m{0.30\linewidth} >{\centering\arraybackslash}m{0.23\linewidth}}
			\hline
			Method   & $AP\uparrow$ & $MR\downarrow$\\
			\hline
			DINO\cite{dino}         & 86.1 & 51.6\\
			MAE\cite{he2022masked}  & 83.7 & 56.5\\
			MAE+DINO\cite{mae+dino} & 86.4 & 50.2\\
			HAP\cite{yuan2023hap}   & 83.3 & 57.5\\
			SOLIDER\cite{solider}   & 85.7 & 51.9\\
			SAIP\cite{saip}         & 87.1 & 49.6\\
			\hline
			ViTKD\cite{vitkd}   & 86.6 & 50.3 \\
			MaskedKD\cite{maskedkd}   & 84.2  & 53.5 \\
			ScaleKD\cite{scalekd}   & 87.4 & 47.5 \\
			TinyMIM\cite{tinymim}   & 86.4 & 50.3 \\
			Proteus\cite{proteus}   & 86.0 & 50.3 \\
			\hline
			DPAL	 & \underline{88.7} & \underline{45.5}\\
			DPAL$^{\star}$ & \textbf{90.2} & \textbf{42.3} \\
			\hline
		\end{tabularx}
	\end{subtable}
	~
	\begin{subtable}[t]{.32\textwidth}
		\caption{Multiple Human Parsing (\%).}\label{table:parsing}
		\begin{tabularx}{\linewidth}{m{0.38\linewidth} >{\centering\arraybackslash}m{0.30\linewidth} >{\centering\arraybackslash}m{0.23\linewidth}}
			\hline
			Method   & $mIoU\uparrow$ & $AP_p\uparrow$ \\
			\hline
			DINO\cite{dino}         & 46.9 & 45.7 \\
			MAE\cite{he2022masked}  & 45.8 & 44.4 \\
			MAE+DINO\cite{mae+dino} & 47.4 & 46.2 \\
			HAP\cite{yuan2023hap}   & 44.1 & 42.8 \\
			SOLIDER\cite{solider}   & 46.8 & 45.7 \\
			SAIP\cite{saip}         & 48.2 & 46.9 \\
			\hline
			ViTKD\cite{vitkd}   & 44.1 & 42.9 \\
			MaskedKD\cite{maskedkd}   & 47.6 & 46.5 \\
			ScaleKD\cite{scalekd}   & 49.0 & 48.3 \\
			TinyMIM\cite{tinymim}   & 47.1 & 46.0 \\
			Proteus\cite{proteus}   & 49.3 & 47.9 \\
			\hline
			DPAL	 & \underline{51.9} & \underline{50.3} \\
			DPAL$^{\star}$ &\textbf{ 55.8} & \textbf{53.3 }\\\hline
		\end{tabularx}
	\end{subtable}
	~
	\begin{subtable}[t]{.33\textwidth}
		\caption{Part-level Attribute Parsing (\%).}\label{table:part-level}
		\begin{tabularx}{\linewidth}{m{0.35\linewidth} >{\centering\arraybackslash}m{0.30\linewidth} >{\centering\arraybackslash}m{0.35\linewidth}}
			\hline
			Method   & \tiny $AP^{box}_{IoU+F_1}$ & \tiny $AP^{segm}_{IoU+F_1}$ \\
			\hline
			DINO\cite{dino}         & 35.4 & 32.9 \\
			MAE\cite{he2022masked}  & 32.0 & 30.3 \\
			MAE+DINO\cite{mae+dino} & 37.7 & 35.2 \\
			HAP\cite{yuan2023hap}   & 33.0 & 30.4 \\
			SOLIDER\cite{solider}   & 36.7 & 34.2 \\
			SAIP\cite{saip}         & 38.0 & 35.5 \\
			\hline
			ViTKD\cite{vitkd}   & 35.6 & 33.4 \\
			MaskedKD\cite{maskedkd}   & 33.3 & 31.0 \\
			ScaleKD\cite{scalekd}   & 37.5 & 34.9  \\
			TinyMIM\cite{tinymim}   & 36.2 & 33.5 \\
			Proteus\cite{proteus}   & 38.0 & 35.4 \\
			\hline
			DPAL	 & \underline{39.8} & \underline{37.0} \\
			DPAL$^{\star}$ & \textbf{42.9} & \textbf{39.8} \\
			\hline
		\end{tabularx}
\end{subtable}
\vspace{1mm}
\begin{subtable}[t]{.33\textwidth}
	\caption{Human Art (\%).}
	\label{table:humanart}
	\begin{tabularx}{\linewidth}{m{0.38\linewidth} >{\centering\arraybackslash}m{0.30\linewidth} >{\centering\arraybackslash}m{0.23\linewidth}}
		\hline
		Method   & $AP\uparrow$ & $AR\uparrow$\\
		\hline
		DINO\cite{dino}         & 65.7 & 69.7\\
		MAE\cite{he2022masked}  & 65.1 & 69.0\\
		MAE+DINO\cite{mae+dino} & 67.4 & 71.1\\
		HAP\cite{yuan2023hap}   & 66.0 & 70.0\\
		SOLIDER\cite{solider}   & 66.7 & 70.6\\
		SAIP\cite{saip}         & 67.5 & 71.2\\
		\hline
		ViTKD\cite{vitkd}   & 68.2 & 71.8 \\
		MaskedKD\cite{maskedkd}   & 64.9 & 68.9 \\
		ScaleKD\cite{scalekd}   & 65.7 & 69.6 \\
		TinyMIM\cite{tinymim}   & 67.3 & 71.0 \\
		Proteus\cite{proteus}   & 69.4 & 73.0\\
		\hline
		DPAL	 & \underline{69.9} & \underline{73.4}\\
		DPAL$^{\star}$ & \textbf{72.9} & \textbf{76.1} \\
		\hline
	\end{tabularx}
\end{subtable}
~
\begin{subtable}[t]{.31\textwidth}
	\caption{Chimpanzee Pose (\%).}
	\label{table:chimpact}
	\begin{tabularx}{\linewidth}{m{0.38\linewidth} >{\centering\arraybackslash}m{0.30\linewidth} >{\centering\arraybackslash}m{0.23\linewidth}}
		\hline
		Method   & $AP\uparrow$ & $AR\uparrow$ \\
		\hline
		DINO\cite{dino}         & 16.1 & 19.2 \\
		MAE\cite{he2022masked}  & 13.9 & 16.7 \\
		MAE+DINO\cite{mae+dino} & 18.2 & 21.2 \\
		HAP\cite{yuan2023hap}   & 13.2 & 15.8 \\
		SOLIDER\cite{solider}   & 16.0 & 18.9 \\
		SAIP\cite{saip}         & 18.2 & 21.3 \\
		\hline
		ViTKD\cite{vitkd}   & 18.9 & 22.4 \\
		MaskedKD\cite{maskedkd}   & 16.2 & 19.1 \\
		ScaleKD\cite{scalekd}   & 20.8 & 24.3 \\
		TinyMIM\cite{tinymim}   & 17.5 & 20.5 \\
		Proteus\cite{proteus}   & 19.7 & 23.2 \\
		\hline
		DPAL	 & \underline{21.9} & \underline{25.5} \\
		DPAL$^{\star}$ &\textbf{ 25.2} & \textbf{29.4 }\\\hline
	\end{tabularx}
\end{subtable}
~
\begin{subtable}[t]{.32\textwidth}
	\caption{Animal Pose (\%).}
	\label{table:ap-10k}
	\begin{tabularx}{\linewidth}{m{0.37\linewidth} >{\centering\arraybackslash}m{0.30\linewidth} >{\centering\arraybackslash}m{0.35\linewidth}}
		\hline
		Method   & $AP\uparrow$ & $AR\uparrow$ \\
		\hline
		DINO\cite{dino}         & 58.0 & 61.6 \\
		MAE\cite{he2022masked}  & 48.8 & 53.5 \\
		MAE+DINO\cite{mae+dino} & 59.5 & 63.6 \\
		HAP\cite{yuan2023hap}   & 50.5 & 54.7 \\
		SOLIDER\cite{solider}   & 57.4 & 61.4 \\
		SAIP\cite{saip}         & 60.8 & 64.6 \\
		\hline
		ViTKD\cite{vitkd}   & 44.0 & 48.7 \\
		MaskedKD\cite{maskedkd}   & 55.3 & 59.0  \\
		ScaleKD\cite{scalekd}   &  62.9  & 66.4  \\
		TinyMIM\cite{tinymim}   & 61.6 & 65.1 \\
		Proteus\cite{proteus}   & 64.7 & 68/1 \\
		\hline
		DPAL	 & \underline{67.0} & \underline{70.3} \\
		DPAL$^{\star}$ & \textbf{69.4} & \textbf{73.0} \\
		\hline
	\end{tabularx}
\end{subtable}
}
\end{table}
\subsection{Experimental Settings}
\textbf{Datasets.} Unless otherwise stated, all models in this paper are pretrained on LUP1M, a subset of 1 million single-person images sampled randomly from LUPerson dataset \cite{luperson}. Specificaly, the multi-persons are synthesized by applying a simple copy-paste technique\cite{ghiasi2021simple} to different resolution images. For downstream task evaluation, the pretrained models are exhaustively evaluated on standard benchmark datasets to ensure comprehensive performance assessment.
Specifically, we adopt Market1501\cite{market1501} and MSMT17\cite{msmt17} for image-to-image ReID (I2I ReID), CUHK-PEDES\cite{cuhk03} and ICFG-PEDES\cite{icfg} for text-to-image ReID (T2I ReID), PA-100K\cite{pa-100k} and PETA\cite{peta} for attribute recognition, 
COCO-Keypoint\cite{coco} for pose estimation, Whole-body COCO\cite{Wholebodycoco} for landmark detection, LIP\cite{lip} for human parsing, 
CrowdHuman\cite{crowdhuman} for pedestrain detection, CIHP\cite{cihp} for multiple human parsing, Fashionpedia\cite{fashionpedia} for part-level attribute parsing. In addition to general human-centric visual perception tasks, we further evaluate generalizability of pretrained models using a cross-domain setting, where models are pretrained using natural person images, and fine-tuned on cross-style or cross-species visual perception tasks using images from unseen domains. Therefore, we consider three representative datasets: 1) Humanart~\cite{dataset:humanart} with person images in unseen styles such as cartoons and sketches; 2) Chimpact-Pose~\cite{dataset:chimpact} with chimpanzee images, and 3) AP-10K~\cite{dataset:ap-10k} with common animal images.

\textbf{Evaluation Metric.} Following previous works, we adopt \(Rank1\) for I2I ReID and T2I ReID, mean accuracy~(\(mAP\)) for attribute recognition, average precision~(\(AP\)) and recall~(\(AR\)) for pose estimation and landmark detection, mean intersection of union~(\(mIoU\)) and mean pixel accuracy~(\(mAcc\)) for human parsing, \(AP\) and missing rate~(\(MR\)) for pedestrain detection, \(mIoU\) and \(AP_p\) for multiple human parsing, \(AP^{box}_{IoU+F_1}\) and \(AP^{segm}_{IoU+F_1}\) for part-level attribute parsing.

\subsection{Implementation Details}
\textbf{Pretraining.} All lightweight HVMs are pretrained or finetuned using 8 A6000 48G GPUs. We employ the AdamW optimizer\cite{adamw} with an effective batch size of 2048 (i.e., 256 per GPU). By default, each model is pretrained from scratch for 100 epochs, initialized with a learning rate of 2.5e-4 and decayed via Cosine Annealing scheduler\cite{loshchilov2016sgdr}. Other settings follow the configuration of DINO\cite{dino}.

\textbf{Finetuning.} To fairly evaluate the pretrained HVMs, any decoder or alignment module (e.g., P-DaDe) is discarded and only pretrained backbone model using different pretraining paradigms (self-supervised/distillation-based) is retained for downstream tasks. For additional details, we refer readers to \textbf{Appendix}.

\subsection{Comparison With State-Of-The-Arts}
In this section, we compare DPAL with existing pretraining methods across a wide range of downstream tasks. Specifically, we conduct a comprehensive evaluation against both self-supervised pretraining paradigms (e.g., DINO \cite{dino}, MAE \cite{he2022masked}) and distillation-based pretraining paradigms (e.g., TinyMIM \cite{tinymim}, Proteus \cite{proteus}). We adopt PATH \cite{tang2023humanbench} as the teacher model and use ViT-Tiny and Swin-Tiny~\cite{liu2021swin} as the student model, training them for 100 epochs on the LUP1M dataset. 

\textbf{Single-person Discriminative Tasks.} We evaluate DPAL's performance on single-person discriminative tasks, including T2I ReID, I2I ReID, and attribute recognition. 
As shown in Table~\ref{tab:comparison with sota}~(a-c), DPAL outperforms SAIP by significant margins of 1.6\% and 8.7\% on Market-1501 and MSMT17 for I2I ReID, and by 5.1\% and 9.9\% on CUHK and ICFG for T2I ReID. For attribute recognition, DPAL outperforms TinyMIM by 0.7\% and 1.5\% on PA100K and PETAzs, respectively. 
Additionally, when using Swin-Tiny as the student backbone, the performance improves by an average of 1.63\% compared to DPAL.
The performance of DPAL on a wide range of single-person discriminative tasks demonstrates that our method is able to effectively learn the global identity patterns from teacher while maintaining strong generalization ability.

\textbf{Dense Prediction Tasks.} We compare DPAL with existing methods on three dense prediction tasks, as shown in Table~\ref{tab:comparison with sota}~(d-f). For pose estimation, DPAL achieved 72.6\% AP and 75.8\% AR, outperforming TinyMiM which scores 70.2\% and 73.6\%, respectively.
In landmark detection, DPAL achieved 48.8\% AP and 61.5\% AR, surpassing TinyMiM by 4.6\% and 4\%.
For human parsing, DPAL achieves $mIoU$ of 55.9, exceeding TinyMiM by 2.9\%. It also achieves 66.7\% mean accuracy ($mAcc$), outperforming Proteus by 2.8\%. 
Most notably, employing Swin-Tiny as the student model demonstrates an average 3.42\% performance improvement over DPAL.
The results substantiate that DPAL successfully learns both global identity patterns and local shape patterns.

\textbf{Multi-person Visual Understanding Tasks.} We evaluate DPAL's performance on three multi-person visual understanding tasks, as shown in Table~\ref{tab:comparison with sota_interaction}~(a-c), In pedestrian detection, DPAL achieves an AP of 88.7\% and an MR of 45.5\%, outperforming SAIP (2nd best) by 1.6\% and 4.1\%, respectively.
In multiple human parsing, DPAL achieves a $mIoU$ of 51.9 and an $AP_p$ of 50.3\%, surpassing Proteus by 2.6\% and 2.4\%, respectively.
Specifically, using Swin-Tiny as the student model yields an average performance gain of 2.92\% relative to DPAL.
In part-level attribute parsing, DPAL delivers the best performance across key evaluation metrics.

\noindent\textbf{Cross-domain Generalization.} As shown in Table~\ref{tab:comparison with sota_interaction}~(d-f), the DPAL consistently outperforms previous self-supervised pretraining/distillation-based pretraining methods on two different scenarios unseen in the pretraining phase, including 1 cross-style recognition task and 2 cross-species recognition tasks. This indicates that learning three typical human visual patterns from large HVMs enables generalizable representations for cross-domain adaptation.

\begin{figure}[h]
  \centering
  \includegraphics[width=\linewidth]{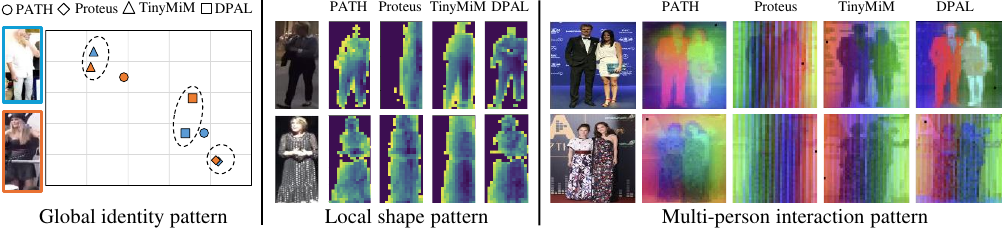}
  \caption{Qualitative comparison between proposed DPAL and state-of-the-art methods.}
  \label{fig:visu}
\end{figure}
\textbf{Qualitative Comparison.} To validate whether DPAL correctly captures the three visual patterns, we visualize the model's outputs. As shown in the figure~\ref{fig:visu}, for global identity patterns, DPAL can effectively distinguish between different instances in the representation space, while both Proteus and TinyMIM struggle to do so. For local shape patterns, DPAL achieves results comparable to those of the teacher model. Additionally, DPAL can successfully differentiate between distinct instances in multi-person images. 

\subsection{Ablation Study} \label{Ablation study}
In this section, we analyze the effectiveness of our learning strategy and seek to best practice for distillation-based pretraining. We adopt I2I ReID, human parsing and pedestrian detection as representative tasks for investigation. For I2I ReID, we evaluate $Rank1$ on Market1501. For human parsing, we evaluate $mIoU$ on LIP. For pedestrian detection, we evaluate $AP$ on CrowdHuman.

\begin{table}[ht]
  \centering
  \fontsize{9pt}{10pt}\selectfont
  \setlength{\tabcolsep}{5pt}  
  \renewcommand{\arraystretch}{1.2}  
  \caption{Comparison of different distillation strategies~(\%). 
  ViT-Ti/16 models are distilled from PATH on LUP1M for 100 epochs.}
  \label{tab:ablation_strategy}
  \begin{tabular}{cccccc}
    \toprule
    $\ell_{g}$&$\ell_{l}$&$\ell_{r}$&I2I ReID& Human parsing &Detection \\ 
    \midrule
    \checkmark &            &            & 95.3 & 52.7 & 87.3  \\
               & \checkmark &            & 93.1 & 55.7 & 88.4  \\
               &            & \checkmark & 92.5 & 53.0 & 86.4  \\
    \checkmark & \checkmark &            & 94.9 & 55.7 & 88.1  \\
               & \checkmark & \checkmark & 93.9 & 55.5 & 88.1  \\
    \checkmark &            & \checkmark & \textbf{95.3} & 53.1 & 86.7  \\
    \checkmark & \checkmark & \checkmark & 95.2 & \textbf{55.9} & \textbf{88.7}  \\
  \bottomrule
  \end{tabular}
\end{table}

\textbf{Distillation Strategies.} To investigate the most suitable distillation strategy, we conduct ablation experiments on three types of learning objectives. We use PATH\cite{tang2023humanbench} as the teacher model and distill it for 100 epochs on the LUP1M dataset. Comparative results are reported in Table~\ref{tab:ablation_strategy}. We find that $\ell_{g}$ focuses more on global information, which allows it to outperform both $\ell_{l}$ and $\ell_{r}$ in single-person discrimination task (95.3\% vs 93.1\% vs 92.5\%). 
As for dense prediction task, $\ell_{l}$ outperforms $\ell_{g}$ and $\ell_{r}$ for human parsing (55.7\% vs 52.7\% vs 53.0\%). 
When three strategies are combined, they achieve strongest generalization ability across three types of downstream tasks. Therefore, we adopt three-level alignment objectives as our default distillation approach.

\begin{table}[ht]
	\centering
	\fontsize{9pt}{10pt}\selectfont
	\setlength{\tabcolsep}{3pt} 
	\renewcommand{\arraystretch}{1.2}  
	\caption{Ablation study on dynamic pattern decoder (\%). Aligning ViT-Ti/16 with PATH-B using or not using dynamic pattern decoder.}
	\label{tab:ablation_d-pade}
	\begin{tabular}{c|cc|cc|cc}
		\toprule
		Setting &Market/MSMT& PA-100K/PETAzs & COCO &  LIP & CrowdHuman & Fashionpedia \\ 
		\midrule
		w/o D-PaDe    & 95.2/83.1 & 65.9/72.4 & 70.9/74.1 &55.2/67.6  & 87.2/49.4 & 38.0/35.4 \\
		w D-PaDe & 95.2/84.3 & 64.3/74.0 & 72.6/75.8 & 55.9/66.7   & 88.7/45.5 & 39.8/37.0 \\
		\bottomrule
	\end{tabular}
\end{table}
\textbf{The Impact of Dynamic Pattern Decoder.} Directly learning different patterns leads to different pixel representations, bringing inter-pattern conflict problem. Experimental results listed in Table~\ref{tab:ablation_d-pade} show that DPAL without D-PaDe brings sub-optimal performance. This demonstrates proposed D-PaDe is capable of alleviating the adverse effect of inter-pattern conflict problem.

\begin{wrapfigure}[17]{r}{0.5\textwidth}
	\centering
	\includegraphics[width=0.48\textwidth]{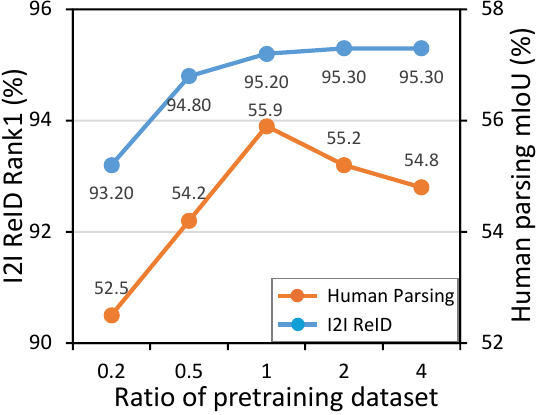}
	\caption{The impact of different scale of datasets on I2I Reid and human parsing tasks.}
	\label{fig:scale_dataset}
\end{wrapfigure}

\textbf{Scale of the Dataset.} To explore the optimal scale of the dataset, we construct five subsets of varying scales (0.2M, 0.5M, 1M, 2M, and 4M samples) from the LUPerson dataset for pretraining. As shown in Figure~\ref{fig:scale_dataset}, we observe that the performance on the 0.2M and 0.5M subsets is significantly worse than on the 1M subset. Moreover, increasing the dataset size does not lead to further performance improvement. Therefore, LUP1M, as the subset of LUPerson, is sufficient to support distillation-based pretraining.

\textbf{Pretraining on Different Teachers.} As shown in Table~\ref{tab:ablation_teacher}, we explore the effects of DPAL on different teacher models. Specifically, we select three teacher models and perform three-level alignment objectives on LUP1M. Notably, DPAL, utilizing a smaller backbone (ViT-Ti/16) and fewer training images (1.2M), achieves superior human parsing $mIoU$ compared to MAE and HAP, with improvements of approximately 12.4\% and 11.5\%, respectively. Moreover, DPAL surpasses PATH in the I2I ReID task by 1.7\%, despite PATH employing a larger model (ViT-B/16) and a significantly larger dataset (12M images). These results demonstrate that DPAL attains enhanced performance while maintaining a more compact model size and reduces training data requirements, indicating its efficiency and strong generalization capability.

\begin{table}[ht]
    \fontsize{9pt}{10pt}\selectfont
    \renewcommand{\arraystretch}{1.2}  
    \caption{Ablation study on different teacher (\%). ViT-Ti/16 models are pretrained with three-level alignment objectives from different teachers on LUP1M for 100 epochs.}
    \label{tab:ablation_teacher}
    \centering
    \begin{tabular}{llllccc}
    \toprule
    Method & Arch & \# imgs & Teacher & I2I ReID & Human Parsing & Detection \\
    \midrule
    HAP-B\cite{yuan2023hap} & ViT-B/16 & 2.1M & - & 95.5 & 54.8 & 89.6 \\
    DPAL & ViT-Ti/16 & 1.2M & HAP-B & 94.1 & 56.1 & 87.7 \\ \midrule
    PATH-B\cite{tang2023humanbench} & ViT-B/16 & 12M & - & 93.5 & 59.1 & 90.1 \\
    DPAL & ViT-Ti/16 & 1.2M & PATH-B & 95.2 & 55.9 & 88.7 \\
    \midrule
    Sapiens-L\cite{khirodkar2024sapiens}& ViT-L/16 & 300M & -  & 89.4 & 34.8 & 89.5\\
    DPAL & ViT-Ti/16 & 1.2M & Sapiens-L & 85.9 & 48.6 & 85.2 \\ 
    
    \bottomrule
    \end{tabular}
\end{table}

\section{Conclusion} \label{Conclusion}
In this paper, we propose DPAL, a novel distillation-based pretraining framework for distilling knowledge of diverse visual patterns from large HVMs into lightweight HVMs without requiring access to inaccessible or strictly constrained pretraining datasets. Specifically, we design a dynamic pattern decoder that 
adaptively extracts three typical human visual patterns conditioning on input image and pattern queries. Finally, we introduce three-level alignment objectives to maximize the effectiveness of knowledge transfer from the teacher to the student. 
Extensive experiments on 12 datasets demonstrate that DPAL achieves strong generalization comparable to much larger models while significantly outperforming previous pretraining methods. Notably, DPAL trains on a limited, unlabeled dataset unseen by large HVMs, providing a more accessible and cost-effective approach to developing generalizable models.


\appendix

\section{Discussion}
\paragraph{Limitation.} The performance of the student modle is influenced by the teacher model used for pretraining. Additionally, due to limitations in computational resource, we have only tested our method on image datasets and downstream tasks. However, our approach is also applicable to human-centric video understanding tasks, which will be explored in our future work.
\paragraph{Broader Impact.} As demonstrated in Section~4, our method outperforms existing pretraining methods across various downstream tasks, highlighting the potential of DPAL as a novel distillation-based pretraining paradigm. 
Moreover, DPAL serves as an efficient knowledge distillation technique that enables the development of compact variants of large HVMs, making them suitable for deployment on resource-constrained edge devices.
Additionally, DPAL eliminates the necessity of accessing the teacher model’s original pretraining datasets by utilizing a relatively small open-source dataset of approximately 1 million images for pretraining.
This pretraining paradigm significantly reduces training costs and enhances the accessibility of DPAL for the research community, thereby broadening its potential applications. Furthermore, the codebase developed in this work is publicly released to promote reproducibility and further advancements in research.

\section{Implementation Details}
\subsection{Model architecture}
\paragraph{Backbone.} We conducted experiments on various student backbones and teacher backbones, with the corresponding settings presented in Table~\ref{tab:architectures}.
\begin{table}[ht]
	\centering
	\fontsize{9pt}{10pt}\selectfont
	\renewcommand{\arraystretch}{1.2}  
	\caption{Configuration of neural architectures. Both Vision Transformer (ViT-X) and Swin Transformer (Swin-X) are used for investigation.} \label{tab:architectures}
	\begin{subtable}[t]{0.8\textwidth}
		\centering
		\label{subtab:vit}
		\begin{tabularx}{\textwidth}{CCCCC}
			\toprule
			Arch&Patch size&Embed dim&Heads&Blocks \\ 
			\midrule
			ViT-Ti & 16 & 192 & 6 & 12 \\
			ViT-S & 16 & 384 & 6 & 12 \\
			ViT-B & 16 & 768 & 12 & 24 \\
			\bottomrule
		\end{tabularx}
	\end{subtable} 
	\\[0.5em]
	\begin{subtable}[t]{0.8\textwidth}
		\centering
		\label{subtab:swin}
		\begin{tabularx}{\textwidth}{CCCCCC}
			\toprule
			Arch&Patch size&Window size&Embed dim&Heads&Blocks \\ 
			\midrule
			Swin-Ti & 4 & 7 & 96 & (3,6,12,24) & (2,2,  6,2) \\
			Swin-S & 4 & 7 & 96 & (3,6,12,24) & (2,2,18,2) \\
			Swin-B & 4 & 7 & 128 & (4,8,16,32) & (2,2,18,2) \\
			\bottomrule
		\end{tabularx}
	\end{subtable} 
\end{table}

\paragraph{Dynamic Pattern Decoder.} The dynamic pattern decoder comprises a self-attention module, a router module, three experts and a dynamic expert generator. The router is responsible for assigning experts to different visual tokens. The experts specialize in handing specific patterns. The dynamic expert generator produces weights for experts conditioned on the visual tokens and pattern queries.

\paragraph{Dynamic Expert Generator.} The dynamic expert generator consists of self-attention, cross-attention, and FFN modules. We design three learnable expert tokens \(T_e=[T_e^1, T_e^2,T_e^3]\), which pass through self-attention, cross-attention, and FFN modules to update the parameters of the three experts. In the cross-attention module, representations from backbone are used as the keys and values, while the expert tokens serve as the queries. The cross-attention module ensures that the parameters of the experts are updated based on the corresponding representations and pattern queries, enabling each expert to selectively focus on the most relevant pattern. 
\paragraph{Router.} We designed a router that dynamically assigns experts to distanct visual patterns, thereby decoupling the alignment learning of three visual patterns. We designed a learnable router token \(T_r\), using the representations extracted by the backbone as keys and values. The routing token dynamically adjusts the weights of different experts \(W_e\) based on different patterns, enabling the model to effectively capture diverse patterns and enhance its performance in complex visual tasks.

\begin{wraptable}[11]{r}{0.45\textwidth} 
	\centering
	\fontsize{9pt}{10pt}\selectfont
	\setlength{\tabcolsep}{5pt}  
	\renewcommand{\arraystretch}{1.2}  
	\caption{Configurations of pretraining.}
	\label{tab:pretraining_details}
	\begin{tabular}{c|c}
		\toprule
		Configuration&Value \\ 
		\midrule
		Batch size & 2048 \\
		Optimizer & AdamW \\
		Learning rate & 2.5e--4 \\
		Learning rate decay & Consine scheduler \\
		Weight decay & 0.05 \\
		Warmup epochs & 10 \\
		Epochs & 100 \\
		Image size & $256\times128$ \\
		\bottomrule
	\end{tabular}
\end{wraptable} 
\subsection{Pretraining Details}
Table~\ref{tab:pretraining_details} shows the default pretraining settings. We initialize transformer blocks with Xavier uniform. We set the batch size to 256 per GPU, using a total of 8 A6000 GPUs for training over 100 epochs. The learning rate linearly warms up during the first 10 epochs and is subsequently decayed according to a cosine decay schedule. The single-person image size is set to $256\times128$ while the multi-person image is set to $256\times256$.

\subsection{Finetuning Details}
We utilize representative methods from downstream tasks as baselines, subsequently replacing their backbones with our pretrained backbones for finetuning. The list of codebases used for evaluation is presented in Table~\ref{tab:baselines}.
\begin{table}[ht]
	\centering
	\fontsize{9pt}{10pt}\selectfont
	\setlength{\tabcolsep}{3pt}  
	\renewcommand{\arraystretch}{1.2}  
	\caption{Implementation codebases and configurations of fine-tuning on 12 datasets. }
	\label{tab:baselines}
	\begin{tabular}{lllccc}
		\toprule
		Task&Dataset&Codebases&Image size &Learning rate&Epoch \\ 
		\midrule
		\multirow{2}{*}{I2I ReID} & Market1501\cite{market1501} & \multirow{2}{*}{SOLIDER~\cite{solider}} & \multirow{2}{*}{$256\times128$} & \multirow{2}{*}{2e-4} & \multirow{2}{*}{120} \\
		~ & MSMT17\cite{msmt17} & ~ & ~ & ~ & ~ \\
		
		\midrule
		\multirow{2}{*}{T2I ReID} & CUHK-PEDES\cite{cuhk03} & \multirow{2}{*}{IRRA~\cite{ds_t2ireid_irra}} & \multirow{2}{*}{$384\times128$} & \multirow{2}{*}{1e-4} & \multirow{2}{*}{60} \\
		~ & ICFG-PEDES\cite{icfg} & ~ & ~ & ~ & ~ \\
		
		\midrule
		\multirow{2}{*}{Attribute recognition} & PA100\cite{pa-100k} & \multirow{2}{*}{SOLIDER~\cite{solider}} & \multirow{2}{*}{$256\times128$} & \multirow{2}{*}{1e-4} & \multirow{2}{*}{25} \\
		~ & PETAzs\cite{peta} & ~ & ~ & ~ & ~ \\
		
		\midrule
		Pose estimation & COCO keypoint\cite{coco} & ViTPose~\cite{pose2} & $256\times192$ & 5e-4 & 210 \\
		
		\midrule
		Landmark detection & Whole-body COCO\cite{Wholebodycoco} & ViTPose~\cite{pose2} & $256\times192$ & 5e-4 & 210 \\
		
		\midrule
		Human parsing & LIP~\cite{lip} & SOLIDER\cite{solider} & $576\times384$ & 7e-4 & 150 \\
		
		\midrule
		Pedestrian detection & CrowdHuman\cite{crowdhuman} & CrowdDet~\cite{ds_pdetection_crowddet} & $1400\times 800$ & 2e-4 & 30 \\
		
		\midrule
		Multiple human parsing & CIHP\cite{cihp} & Cpi-parser~\cite{ds_mhp_cpiparser} & $1333\times800$ & 2e-2 & 25 \\
		
		\midrule
		Part-level attribute parsing & Fashionpedia\cite{fashionpedia} & KE-RCNN~\cite{ds_part_parse_kercnn} & $1024\times1024$ & 1e-4 & 32 \\
		
		\bottomrule
	\end{tabular}
\end{table}

\section{Ablation Study}

\begin{table}[ht]
	\centering
	\fontsize{9pt}{10pt}\selectfont
	\setlength{\tabcolsep}{10pt} 
	\renewcommand{\arraystretch}{1.2}  
	\caption{Ablation study on variants of pattern decoder (\%). Aligning ViT-Ti/16 with PATH-B using MAE-style decoder, Standard MoE or D-PaDe.}
	\label{tab:ablation_variant_decoder}
	\begin{tabular}{c|ccc}
		\toprule
		Setting & I2I ReID & Human Parsing & Detection \\ 
		\midrule
		MAE-style    & 95.1 &55.0  & 87.8 \\
		Standard MoE  & 94.1 &55.9  & 88.5 \\
		D-PaDe & 95.2 &  55.9  & 88.7\\
		\bottomrule
	\end{tabular}
\end{table}
\subsection{The Study of Variants of Pattern Decoder} 
This section study the variants of pattern decoder, including MAE-Style~\cite{he2022masked} decoder using two transformer blocks, Standard MoE and proposed D-PaDe. As shown in Table~\ref{tab:ablation_variant_decoder} show that D-PaDe is the best choice for distillation-based pretraining by far. 

\begin{table}[!h]
    \setlength{\tabcolsep}{1.5pt}
    \renewcommand\arraystretch{1.25}
    \caption{Impact of model architecture. We employ PATH-B as teacher model and perform distillation with DPAL on four student architectures.
    }
    \label{tab:Impact_of_Arch}
    { 
      \fontsize{9pt}{10pt}\selectfont
      \renewcommand{\arraystretch}{1.2}  
    \begin{subtable}{\linewidth}
    \caption{Single-person discrimitive tasks (\%).} \label{table:single-person dis}
    \centering
    \begin{tabularx}{\textwidth}{CCC|CC|CC|CC}
        \toprule
        \multirow{2}{*}{Arch}& \multirow{2}{*}{Type} & \multirow{2}{*}{\#Param} & \multicolumn{2}{c|}{I2I ReID} & \multicolumn{2}{c|}{T2I ReID} & \multicolumn{2}{c}{Attribute recognition} \\ \cline{4-9}
        ~&~&~&Market$\uparrow$&MSMT17$\uparrow$&CUHK$\uparrow$&ICFG$\uparrow$&PA100K$\uparrow$&PETAzs$\uparrow$\\
        \midrule
        ViT-Ti/16 & ViT & 5M & 95.2 & 84.3 & 64.3 & 56.0 & 82.4 & 74.0 \\
        ViT-S/16 & ViT & 21M & 95.8 & 86.1 & 65.8 & 58.5 & 83.9 & 74.1 \\
        Swin-Ti/4 & Hybrid & 27M & 96.4 & 86.2 & 66.9 & 58.5 & 83.1 & 74.9 \\
        Swin-S/4 & Hybrid & 49M & \textbf{96.9} & \textbf{88.2} & \textbf{69.6} & \textbf{60.0} & \textbf{85.9} & \textbf{77.1} \\
        \bottomrule
    \end{tabularx}
    \end{subtable}
    \begin{subtable}{\linewidth}
    \caption{Single-person dense prediction tasks (\%).} \label{table:single-person dense}
    \centering
    \begin{tabularx}{\textwidth}{CCC|CC|CC|CC}
        \toprule
        \multirow{2}{*}{Arch}& \multirow{2}{*}{Type} & \multirow{2}{*}{\#Param} & \multicolumn{2}{c|}{Pose estimation} & \multicolumn{2}{c|}{Landmark detection} & \multicolumn{2}{c}{Human parsing } \\ \cline{4-9}
        ~&~&~&$AP\uparrow$&$AR\uparrow$&$AP\uparrow$&$AR\uparrow$&$mIoU\uparrow$&$mAcc\uparrow$\\
        \midrule
        ViT-Ti/16 & ViT & 5M & 72.6 & 75.8 & 48.8 & 61.5 & 55.9 & 66.7 \\
        ViT-S/16 & ViT & 21M & 73.3 & 76.3 & 53.1 & 65.5 & 58.1 & 68.7 \\
        Swin-Ti/4 & Hybrid & 27M & 75.1 & 78.1 & 53.9 & 65.7 & 59.3 & 69.7 \\
        Swin-S/4 & Hybrid & 49M & \textbf{76.3} & \textbf{79.4} & \textbf{55.6} & \textbf{67.2}	& \textbf{60.7} & \textbf{71.5} \\
        \bottomrule
    \end{tabularx}
    \end{subtable}
    \begin{subtable}{\linewidth}
    \caption{Multi-person visual understanding tasks (\%).} \label{table:multi-person}
    \centering
    \begin{tabularx}{\textwidth}{CCC|CC|CC|CC}
        \toprule
        \multirow{2}{*}{Arch}& \multirow{2}{*}{Type} & \multirow{2}{*}{\#Param} & \multicolumn{2}{c|}{Pedestrian detection} & \multicolumn{2}{c|}{Multiple human parsing} & \multicolumn{2}{c}{Part-level attribute parsing } \\ \cline{4-9} 
        ~&~&~&$AP\uparrow$&$MR\downarrow$&$mIoU\uparrow$&$AP_p\uparrow$&\tiny $AP^{box}_{IoU+F_1}\uparrow$&\tiny $AP^{segm}_{IoU+F_1}\uparrow$\\
        \midrule
        ViT-Ti/16 & ViT & 5M & 88.7 & 45.5 & 51.9 & 50.3 & 39.8 & 37.0 \\
        ViT-S/16 & ViT & 21M & 89.2 & 42.9 & \textbf{55.9} & \textbf{53.4} & 42.9 & 39.3 \\
        Swin-Ti/4 & Hybrid & 27M & \textbf{90.2} & \textbf{42.3} & 55.8 & 53.3& 42.9 & 39.8 \\
        Swin-S/4 & Hybrid & 49M & 89.7 & 43.1 & 55.2 & 52.4 & \textbf{44.9} & \textbf{41.1} \\
        \bottomrule
    \end{tabularx}
    \end{subtable}
    \begin{subtable}{\linewidth}
    	\caption{Cross-domain perception tasks (\%).} \label{table:cross-domain}
    	\centering
    	\begin{tabularx}{\textwidth}{CCC|CC|CC|CC}
    		\toprule
    		\multirow{2}{*}{Arch}& \multirow{2}{*}{Type} & \multirow{2}{*}{\#Param} & \multicolumn{2}{c|}{Humanart} & \multicolumn{2}{c|}{Chimpact-Pose} & \multicolumn{2}{c}{AP-10K } \\ \cline{4-9}
    		~&~&~&$AP\uparrow$&$AR\uparrow$&$AP\uparrow$&$AR\uparrow$&$AP\uparrow$&$AR\uparrow$\\
    		\midrule
    		ViT-Ti/16 & ViT & 5M & 69.9 & 73.4 & 21.9 & 25.5 & 67.0 & 70.3 \\
    		ViT-S/16 & ViT & 21M & 72.0 & 75.5 & 24.7 & 28.4 & 69.0 & 72.4 \\
    		Swin-Ti/4 & Hybrid & 27M & 72.9 & 76.1 & 25.2 & 29.4 & 69.4 & 73.0 \\
    		Swin-S/4 & Hybrid & 49M & \textbf{75.1} & \textbf{78.3} & \textbf{27.8} & \textbf{32.1}	& \textbf{71.5} & \textbf{74.7} \\
    		\bottomrule
    	\end{tabularx}
    \end{subtable}
}
\end{table}
\subsection{Influence of Model Architecture}
We evaluate the performance of DPAL on downstream tasks with different model architectures. We use PATH-B as the teacher model and perform distillation for 100 epochs by default. As shown in the Table~\ref{tab:Impact_of_Arch}, we employ vision transformer~\cite{vit} for the ViT architecture and swin transformer~\cite{swin} for the hybrid architecture. The other settings are the same as in Section~4.1 ans Section~4.2.
We observe a consistent improvement in model performance concomitant with the increasing model parameters, as exemplified by I2I ReID task where $Rank1$ increases by 0.6\%~(+16M), 1.2\%~(+22M), 1.7\%~(+44M) compared to ViT-Tiny. However, this improvement is accompanied by a corresponding increase in training costs.
Moreover, our method is model-agnostic, demonstrating strong performance on both ViT and hybrid architectures.

\subsection{Influence of Teacher Size}
We investigate whether employing teacher models with larger size enhances the performance of the student model. Specifically, we employ PATH-B and PATH-L as teacher models to distill ViT-Tiny. 
The results presented in Table~\ref{tab:teacher_size} indicate that increasing the size of the teacher model does not yield performance gains across a wide range of downsream tasks. This may be due to the larger gap between the larger teacher models and the student model, which is also mentioned in the TinyMIM\cite{tinymim}.
\begin{table}[!h]
    \setlength{\tabcolsep}{1.5pt}
    \renewcommand\arraystretch{1.25}
    \caption{Impact of teacher size. We use ViT-Tiny as the student model and perform DPAL distillation separately with teacher models of two different sizes.}
    \label{tab:teacher_size}
    { 
      \fontsize{9pt}{10pt}\selectfont
      \setlength{\tabcolsep}{2pt}  
      \renewcommand{\arraystretch}{1.2}  
    \begin{subtable}{\linewidth}
    \caption{Single-person discrimitive tasks (\%).} \label{table:single-person dis}
    \centering
    \begin{tabularx}{\textwidth}{C|CC|CC|CC}
        \toprule
        \multirow{2}{*}{Teacher}& \multicolumn{2}{c|}{I2I ReID} & \multicolumn{2}{c|}{T2I ReID} & \multicolumn{2}{c}{Attribute recognition} \\ \cline{2-7}
        ~&Market$\uparrow$&MSMT17$\uparrow$&CUHK$\uparrow$&ICFG$\uparrow$&PA100K$\uparrow$&PETAzs$\uparrow$\\
        \midrule
        PATH-B & 95.2 & \textbf{84.3} & 64.3 & \textbf{56.0} & 82.4 & 74.0 \\
        PATH-L & \textbf{95.2} & 83.7 & \textbf{66.0} & 55.9 & \textbf{82.7} & \textbf{74.3} \\
        \bottomrule
    \end{tabularx}
    \end{subtable}
    \begin{subtable}{\linewidth}
    \caption{Single-person dense prediction tasks (\%).} \label{table:single-person dense}
    \centering
    \begin{tabularx}{\textwidth}{C|CC|CC|CC}
        \toprule
        \multirow{2}{*}{Teacher}& \multicolumn{2}{c|}{Pose estimation } & \multicolumn{2}{c|}{Landmark detection} & \multicolumn{2}{c}{Human parsing } \\ \cline{2-7}
        ~&$AP\uparrow$&$AR\uparrow$&$AP\uparrow$&$AR\uparrow$&$mIoU\uparrow$&$mAcc\uparrow$\\
        \midrule
        PATH-B & 72.6 & 75.8 & \textbf{48.8} & \textbf{61.5} & \textbf{55.9} & \textbf{66.7} \\
        PATH-L & \textbf{72.7} & \textbf{78.2} & 48.6 & 61.2 & 55.7 & 66.6 \\
        \bottomrule
    \end{tabularx}
    \end{subtable}
}
\end{table}

\begin{figure}[t]
	\centering
	\includegraphics[width=\linewidth]{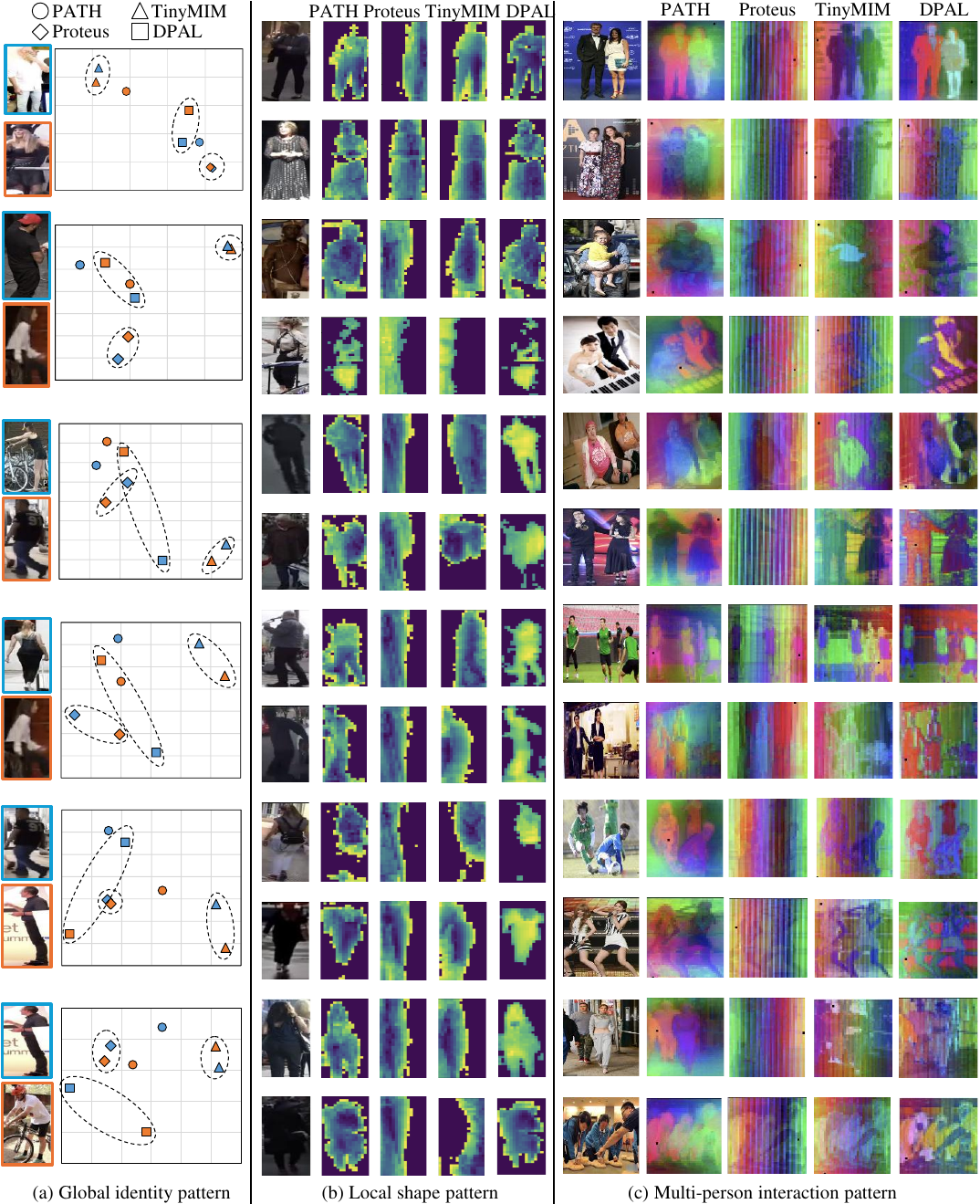}
	\caption{Visualization of learned patterns among four models. From left to right: (a) global identity pattern, (b) local shape pattern, and (c) multi-person interaction pattern. }
	\label{fig:visu2}
\end{figure}

\section{Visualization Results}
We provide additional visualization results in Figure~\ref{fig:visu2}. 
First, we visualize the class token representation space of PATH~\cite{tang2023humanbench}, Proteus\cite{proteus}, TinyMIM\cite{tinymim} and DPAL for two single-person images to investigate the models' ability to learn global identity patterns. As shown in Figure~\ref{fig:visu2}~(a), DPAL and PATH distinctly separate the two instances in the representation space, whereas the other methods do not.
Second, we conduct principal component analysis~(PCA) visualization to investigate the model's capability in capturing local shape patterns. DPAL successfully captures local body shape patterns comparable to those of PATH, while the others fail to capture the whole structure of a person instance. 
Third, we perform PCA visualization on multi-person images. Similar to PATH, DPAL is able to distinguish different individuals that are depicted by different colors in the visualization. This demonstrates that DAPL has successfully enabled lightweight model to acquire multi-person interaction patterns.

\bibliographystyle{ACM-Reference-Format}
\bibliography{ref}

\end{document}